# 基于矩阵填充模型的成绩预测
# Based on Graph-VAE Model to Predict Student's Score

## Abstract

The OECD pointed out that the best way to keep students up to school is to intervene as early as possible [1]. Using education big data and deep learning to predict student's score provides new resources and perspectives for early intervention. Previous forecasting schemes often requires manual filter of features , a large amount of prior knowledge and expert knowledge. Deep learning can automatically extract features without manual intervention to achieve better predictive performance. In this paper, the graph neural network matrix filling model (Graph-VAE) based on deep learning can automatically extract features without a large amount of prior knowledge. The experiment proves that our model is better than the traditional solution in the student's score dataset, and it better describes the correlation and difference between the students and the curriculum, and dimensionality reducing the  vector of coding result is visualized, the clustering effect is consistent with the real data distribution clustering. In addition, we use gradient-based attribution methods to analyze the key factors that influence performance prediction.

## 摘要

经合组织指出想要让学生跟上学业进度，最好的方式是尽早干预[1]，利用教育大数据和深度学习对学生成绩进行预测为尽早干预提供了新的资源和视角。以往的预测方案往往需要手动筛选特征和大量的先验知识以专家知识。深度学习可以不经人工干预，自动提取特征，从而达到更好的预测性能。本文提出基于深度学习的图神经网络矩阵填充模型（Graph-VAE）可以自动提取特征，不需要大量的先验知识。实验证明我们的模型在学生成绩数据集上的效果比传统解决方案好，更好的刻画了学生与课程之间的相关性和差异性，且作为编码结果的向量降维可视化后聚类效果符合真实数据分布聚类。此外，我们利用基于梯度的归因方法分析出影响成绩预测的关键因素。

# 1 Introduction

"数据驱动学校，分析变革教育"的大数据时代已经来临，如何利用大数据的优势、挖掘出其中的模式从而针对性改进教学成为亟需解决的问题。经济合作与发展组织在2018年发布的《2018教育概览》[1]数据调查发现，期末成绩不合格之后再让学生留级并不能达到让学生跟上进度的目的。想要学生跟上学业进度，最好的方式是尽早干预。若能准确预测学生成绩，分析影响学生成绩的因素可以为学校尽早干预学生学习状况、培养方案的改进以及学生选课提供指导意义，并且低分预测可以警示学生及时改正学习中的错误。

现有的成绩预测方法大致分为贝叶斯网络模型[2,3,4]、决策树模型[5,6,7]和典型的

推荐算法[8,9,10,11]等,它们需要手动提取特征、大量的先验知识和专家知识的输入,或者只能预测指定课程成绩或者效果不佳。深度学习是一种新兴的能自动提取特征的算法,其中的卷积神经网络和循环神经网络在处理图像数据及序列化数据领域甚至超过了人类的水平。但是,卷积神经网络和循环神经网络无法接受图结构的数据作为输入,图卷积神经网络(Graph Convolutional Networks,GCN)[12,13]是在卷积神经网络上面的改进,因其可以很好地抓取图结构数据的特征而被广泛使用。

由于学生选课的数据可以分为学生节点、课程节点和选课成绩,更适合建模成二分图形式的图结构数据,本文提出的成绩预测模型(Graph-VAE)利用图卷积神经网络,实现了自动提取特征并且不需要大量的先验知识以及专家知识。该成绩预测模型本质上是一个矩阵填充问题,这一问题与推荐系统中使用矩阵填充模型预测推荐类似。受矩阵填充中 GCMC 模型[14]的启发,结合不同的课程都有各自的特点,不同学生之间也存在差异性以及课程和学生之间存在着某种潜在的相关性,我们提出使用变分自编码器[15]的高斯混合模型刻画其分布,其中图卷积神经网络[12]作为 VAE 的编码部分。该模型将学生节点和课程节点特征信息以及节点间的成绩关系矩阵输入 GCN 进行编码,从而获得学生成绩数据的内在特征,得到节点的嵌入表达经变分重整化再输入解码器,获得与输入邻接矩阵尺寸相同的矩阵——也就是预测的学生成绩。

我们实验对比了经典的推荐算法,验证了模型的有效性。并且,我们还通过可视分析方法观察和分析了模型训练得到的特征向量所蕴含的语义特征,表明我们模型能够学习得到节点隐式的特征。此外,我们利用基于梯度的归因方法[17]分析出影响成绩预测的因素。

综上所述，本文主要贡献如下：

1）提出并实现一种基于无监督学习模型 Graph-VAE，考虑了学生和课程之间的关系；

2）通过大规模实验，对比了其它三类推荐预测成绩方法，验证了该模型的有效性；

3）通过对模型隐藏层学出的特征向量进行可视分析，我们有效的解释了模型学出的特征向量的语义含义；

4）通过对预测结果进行归因分析，有效地挖掘出数据之间的空间相关性。

本文以下章节安排如下：第二章，相关工作；第三章，Graph-VAE 模型介绍；第四章，实验结果；第五章，总结。

## 2 Related Work

当前已有大量研究工作对学生成绩进行预测，但没有研究者从学生用户和课程之间的关系研究它们的空间分布来实现成绩预测功能。黄建明[2]提出的贝叶斯网模型，需要手动筛选特征，并且预测一门课程的成绩；武彤和王秀坤[5]提出的人为确定阈值的决策树模型，需要大量先验知识输入；刘俊岭，李婷等[7]提出利用电子签到系统数据预测成绩，它需要从大量的考勤、作业和课堂位置信息数据中推理出学生心理。

本文提出的成绩预测模型本质上是一个矩阵填充问题，这一问题与推荐系统中使用矩阵填充模型预测推荐类似，因此可以借鉴推荐系统的算法。经典的推荐算法可划分以下几大类：1）协同过滤推荐算法[8]，包括基于邻域的协同过滤（基

于用户和基于项）和基于模型的协同过滤（受限玻尔兹曼机、矩阵因子分解（如奇异值分解，奇异值分解++）、贝叶斯网络、聚类、分类、回归等），这些方法面临冷启动问题，并且很难提供解释；2）基于内容的推荐算法[10]，包括机器学习方法（朴素贝叶斯、支持向量机、决策树等）和信息检索（TF-IDF（Term Frequency–Inverse Document Frequency）算法、BM25（Best Match25）算法），这类方法的项内容必须是机器可读的和有意义的，很难联合多个项的特征；3）混合推荐算法[11]，该类算法综合利用协同过滤推荐算法和基于内容的推荐算法各自的优点同时抵消各自的缺点，算法输入通常使用用户和待推荐物品的内容特性与惯用数据，同时从两种输入类型中获益，该类算法没有冷启动问题，没有流行度偏见，可推荐有罕见性质的项，可以实现多向性，但是需要同时考虑协同过滤推荐算法和基于内容推荐算法，才能得到优缺点的恰当平衡。

相比之下，我们的成绩预测模型主要从学生和课程的空间相关性进行研究，利用高斯混合模型刻画学生和课程之间的相关性以及差异性，不用手动筛选特征，不需要大量先验知识和专家知识并且可以精确的预测得分区间。实验结果表明，我们的成绩预测模型效果优于典型的推荐算法。

## 3 Methods

本章主要从成绩数据的二分图建模、模型的整体架构、编码器和解码器的实现以及变分重整化等五个方面介绍我们的成绩预测模型 Graph-VAE。

### 3.1 二分图结构

本文将学生成绩数据建模成二分图结构，如图 1（a）所示：每个学生/课程都当成一个节点，学生如果选了课程则用无向边连接两个节点，无向边的权重为

学生的课程成绩。这样一个无向二分图可以表示为 $G=(U,V,R)$，$U$ 为学生节点集合，$V$ 为课程节点集合，$R$ 为图 G 的边集。$u_i \in U$（其中 $i \in \{1,\cdots,m\}$）表示第 i 个学生节点，$v_j \in V$（$j \in \{1,\cdots,n\}$）表示第 j 个课程节点，m 为学生总数，n 为课程总数，图的总节点数 $N=m+n$。边的权重 □课程成绩/10□ +1，$r \in \{1,\cdots,10\}$。二分图转化成邻接矩阵 $M = \begin{bmatrix} 0 & A^T \\ A & 0 \end{bmatrix}$，$M \in R^{N \times N}$ 形式存储，如图 1（b）所示，矩阵 $A$ 大小为 m×n，$A_{u_i,v_j} = r$，表示学生 $u_i$ 在课程 $v_j$ 上的成绩为 r，$A^T$ 是 $A$ 的转置矩阵（无向图的邻接矩阵为对称矩阵）。

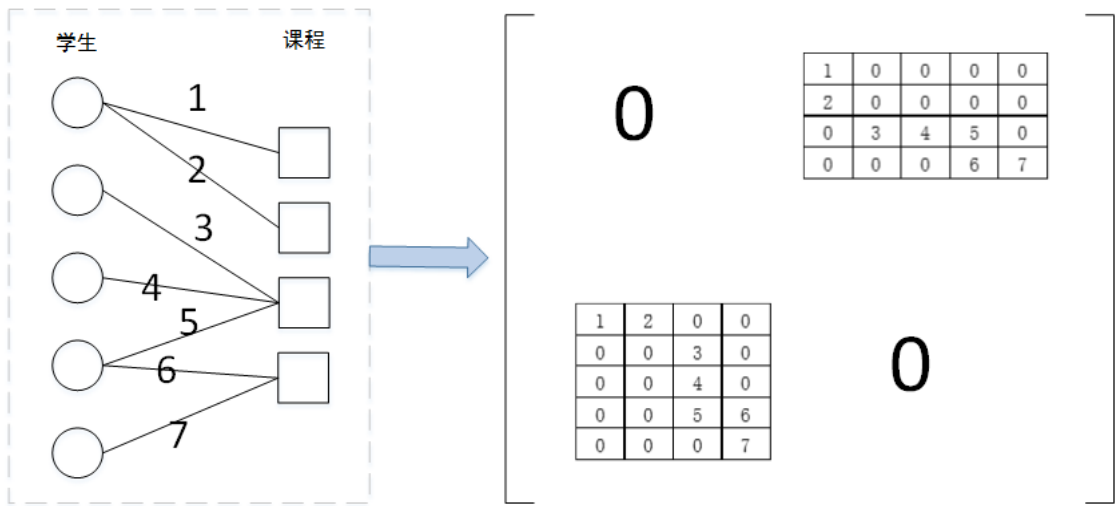

（a）学生-课程二分图　　　　　　（b）邻接矩阵表示
图 1 二分图及其邻接矩阵示意图

将成绩划分成 1-10 共 10 个等级，可以大大减少计算量，且我们假设获得不同成绩等级的学生之间有较大差别，则倾向给出不同的成绩等级的课程间亦然。按照权值 $r$ 将邻接矩阵 M 分解成 10 个等级的 0-1 矩阵 $M_1$-$M_{10}$，$M_r = \begin{bmatrix} 0 & A_r^T \\ A_r & 0 \end{bmatrix}$。$A_r$ 为大小为 m×n 的 0-1 矩阵，$r \in \{1,\cdots,10\}$。若学生 i 所选课程 j 的成绩为 r，则 $A_r$ 第 i 行第 j 列的值为 1，而 $A_{r'}$ 对应的第 i 行第 j 列的值为 0，$r' \in \{1,\cdots,10\}$ 且 $r' \neq r$。

## 3.2 整体架构

本文提出的学生成绩预测模型 Graph-VAE 整体架构如图 2 所示。我们将现有的课程信息、学生信息和成绩数据整合成二分图的形式，将二分图 G 的邻接矩阵 M 和顶点的初始化表示 X 作为 VAE 编码器的输入，经过 GCN 编码之后，输出成绩预测矩阵 $\overline{M}$，完成矩阵填充。Graph-VAE 模型主要包括 encoder 部分和 decoder 部分。输入数据 X 经过 2 层 GCN+ReLu 之后输出编码的均值 Z_mean 和标准差的对数 Z_log_std,通过 N(0,1)采样得到输出 Z,作为 decoder 的输入部分，数据经过 decoder 得到成绩预测矩阵。

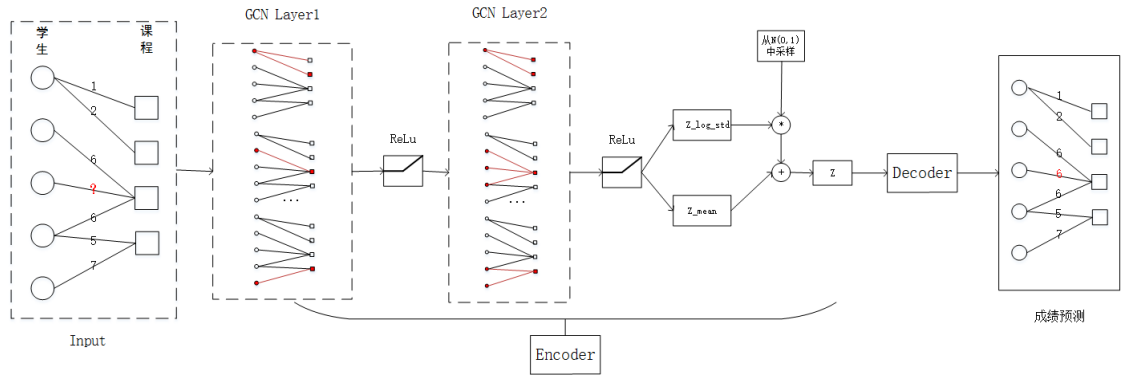

图 2 Graph-VAE 模型工作流程

## 3.3 编码器

图 G 的度矩阵 $D$ 为对角矩阵，$D \in R^{N \times N}$，顶点 $u_i$ 的度 $d(u_i) = \sum_{r=1}^{10} \sum_{j=1}^{n} A_{r,u_i,v_j}$，其中 $A_{r,u_i,v_j}$ 表示学生 $u_i$ 所选的课程 $v_j$ 成绩在 $r$ 等级区间。$W_1, W_2, ..., W_{10}$，$W \in \mathbb{R}^{K \times K}$ 为初始化的权重矩阵，$K$ 为特征的维度数。每个节点 i 的特征描述 $x_i$ 组成的特征矩阵 $X \in R^{N \times K}$，$X \in \mathbb{R}^{N \times K}$ 也为初始化的矩阵。将图 G 的邻接矩阵 M 和初始化的节点矩阵 X 输入 GCN 编码器学习数据之间的潜在规则，经过激活函数 ReLU 得

到隐藏层表示，最终得到学生顶点编码矩阵 $E_u \in \mathbb{R}^{m \times K}$ 和课程顶点编码矩阵 $E_v \in \mathbb{R}^{n \times K}$ 的过程如下：

$$\begin{bmatrix} E_u \\ E_v \end{bmatrix} = encoder(M_1, \ldots, M_{10}) = \text{Re}LU\left(\begin{bmatrix} H_u \\ H_v \end{bmatrix} W\right), \quad \text{（公式 1）}$$

$$\begin{bmatrix} H_u \\ H_v \end{bmatrix} = \text{Re}LU(sum(D^{-1}M_1 X W_1, \ldots, D^{-1}M_{10} X W_{10})) \quad \text{（公式 2）}$$

其中，$H_u \in \mathbb{R}^{N_u \times E}$、$H_v \in \mathbb{R}^{N_v \times E}$ 为中间结果。$D^{-1}M_r X W_r, r \in \{1, \cdots, 10\}$ 相当于综合度矩阵、邻接矩阵包含的信息。邻接矩阵用以引入"按成绩等级分割后"的信息。X 的引入是调节矩阵的尺寸。$W_r$ 用以分别学习 10 个成绩等级的模式，先验假设是不同等级之间的模式也不同，需要分别对待。$Sum(\cdot)$ 加和操作是对各分块的消息进一步传递融合。

模型预测能力体现在随着损失函数的下降，模型的各项参数发生更迭，原图和子图、学生顶点和课程顶点的信息会产生交换。

令：

$$Z = [Z_1^T, \ldots, Z_n^T, Z_{n+1}^T, \ldots, Z_N^T] = \begin{bmatrix} E_u \\ E_v \end{bmatrix}^T 。 \quad \text{（公式 3）}$$

则第 i 名学生选修第 j 门课程可表示为 $(Z_i, Z_{n+j})$，$i=\{1,2,\ldots,n\}$。显然，上述的编码器是可以序贯式堆叠（前一个编码器的输出作为下一个编码器的输入）以达到更大程度抽取数据特征的目的。如图 3 所示，(a)部分表示原始的二分图，其中圆形表示学生节点，矩形表示课程节点，边上的数字表示为成绩。(b)部分表示节点之间的消息传递，其中圆形表示学生节点，矩形表示课程节点，圆角矩形表示神经元，节点右上角表示节点所携带的信息（分别用红、橙、黄、绿、青、蓝、紫颜色的矩形表示节点 A、B、C、D、E、F、G 自身所携带信息）。在卷积前，节点只携带自身信息。第一层卷积时，节点将其携带信息传递到其相邻节点，经

过卷积之后节点学习到其邻居节点的信息。第二层卷积时，节点及其携带的邻居节点的信息传递到其下一跳节点，经过卷积之后节点学习到其两跳节点的信息。要预测节点 A 到节点 F 的成绩，经过 GCN 可以学习到节点 A、B、C 成绩分布相似，从而根据 B、C 到 F 成绩区间预测节点 A 到节点 F 的成绩。

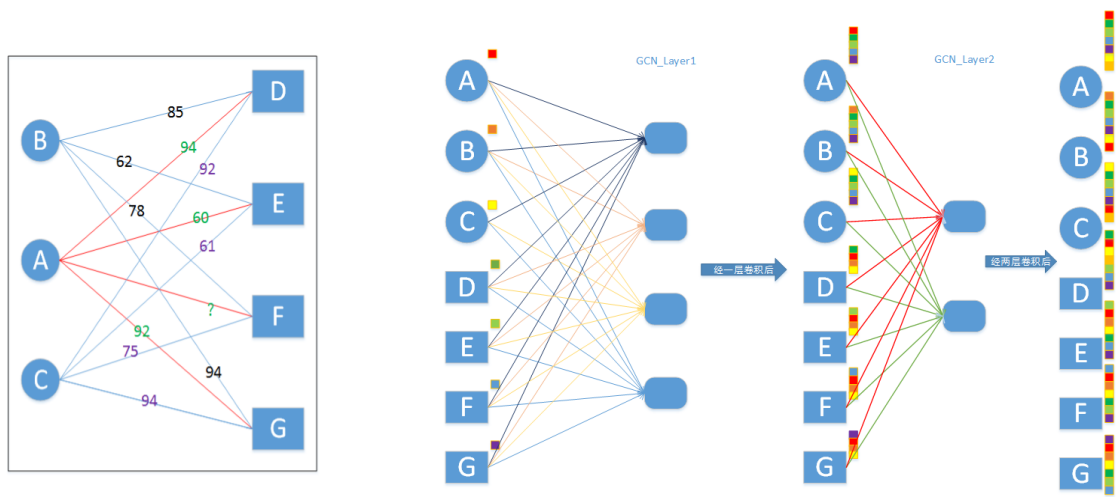

图 3 (a) 二分图数据　　　　图 3 (b) 消息在 GCN 中的传递

图 3　节点消息在 GCN 中的传递

### 3.3 变分重整化

$\mu_i$ 和 $\sigma_i$ 分别表示 $Z_i$ 的均值向量和标准差向量。$\varepsilon_i$ 是从标准正态分布 $N(0,I)$ 取样得到的 E 维向量。

$$Z_i = \mu_i + \varepsilon_i \odot \sigma_i (i = \{1,2,...,N\}) \quad (公式4)$$

$$Z = [Z_1^T,...,Z_{N_u}^T, Z_{N_u+1}^T,...,Z_N^T] \quad (公式5)$$

图 4 所示即为向量 $Z_i$ 的变分重整化示意图，对应于图 2 中 Encoder 的后半部分。由于 VAE 的特性，需要约束 P(Z) 中的每一维都向标准正态分布 $N(0,I)$ 接近，即极小化下式：

$$\cos t_1 = N * KL[P(Z_i) \| \mathcal{N}(0,I)] = -\frac{N}{2}\sum_{i=1}^{E}\left(1+\log\sigma_i^2 - \mu_i^2 - \sigma_i^2\right) \quad (公式6)$$

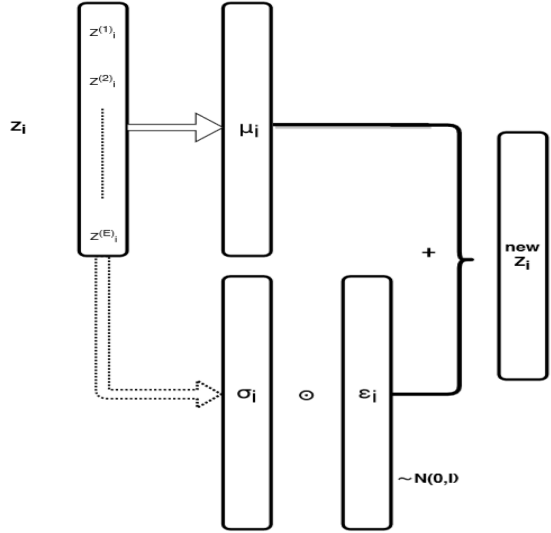

图 4 变分重整化示意图

## 3.4 解码器

VAE 自编码器经过变分重整化输出的 Z 输入解码器，解码器输出 $\overline{M}_r$：

$$Z'_r = (Z \odot H_r) \times Z^T \qquad \text{（公式 7）}$$

其中，$H_r$ 表示 $r$ 通道的权重矩阵，$Z'_r$ 表示第 $r$ 通道的输出。$H_r$ 大小为 $N \times E$。

使用 softmax 将矩阵元素的值"挤压"到 (0,1) 区间内：

$$\overline{M}_{r,ij} = \frac{Z'_{r,ij}}{\sum_{i=1}^{N}\sum_{j=1}^{N}\exp(Z'_{r,ij})} \qquad \text{（公式 8）}$$

$\overline{m}_{r,ij}$ 表示 $\overline{M}_r$ 中第 i 行第 j 列的元素。$Z'_{r,ij}$ 同理。

成绩预测模型的输出 $\overline{M}_r$ 和模型的输入 $M_r$ 为同型矩阵，即为预测对应课程的成绩结果。

## 3.5 优化目标

不同的损失函数，对应 $p(x|z)$ 的不同概率分布假设，我们优化的是 $M$ 的最大似然值，并非利用 $M$ 和 $\overline{M}$ 的差值距离衡量。训练过程，我们最小化负的 $log$ 似然值，并用交叉熵 $cost$ （见公式（9））来度量 $M$ 和 $\overline{M}$ 的差异，$cost$ 越小，$M$ 和 $\overline{M}$ 越接近。

$$\cos t_2 = -\sum_{r=1}^{10}\sum_{j=1}^{N}\sum_{i=1}^{N}\left[M_{r,ij}*\log\left(\overline{M_{r,ij}}\right)+\left(1-M_{r,ij}\right)*\log\left(1-\overline{M_{r,ij}}\right)\right] \quad （公式9）$$

结合变分重整化时的目标函数公式（6），模型 Graph-VAE 的最小化目标函数为 $Loss = \cos t_1 + \cos t_2$：

$$Loss = -\sum_{r=1}^{10}\sum_{j=1}^{N}\sum_{i=1}^{N}\left[M_{r,ij}*\log\left(\overline{M_{r,ij}}\right)+\left(1-M_{r,ij}\right)*\log\left(1-\overline{M_{r,ij}}\right)\right] - \frac{N}{2}\sum_{i=1}^{E}\left(1+\log\sigma_i^2 - \mu_1^2 - \sigma_i^2\right) \quad （公式10）$$

# 4 Experiment Evaluation

### 4.1 数据集

成绩数据集取自中南大学信息院 2010 年至 2016 年 12 月的研究生成绩数据，共有学生顶点数 $N_u = 369$、课程顶点数 $N_v = 142$。由于学校培养方案 4 年变动一次，因此将总数据集分割为三部分：2015 年 9 月-2016 年 7 月（记为 data1）、2012 年 9 月-2016 年 7 月（记为 data2）以及 2010 年 1 月-2016 年 12 月（记为 data3）。

本文使用 RMSE[20]（均方根误差）来评估成绩预测模型的预测误差。RMSE 是均方误差的算术平方根，是最常用来评估预测数据与真实数据差异的一个指标，计算如公式（11）所示。该评价指标反映预测值和真实值的差距，值越接近 0，差距越小。

$$MSE = \frac{1}{10}\sum_{r}^{N=10}|M_r - \overline{M_r}|^2$$
$$RMSE = \sqrt{MSE}$$
公式（11）

### 4.2 基本实验

将 data1、data2、data3 三组数据集各自划分成 75%、10%和 5%作为训练集、测试集、验证集。模型学习率为 0.1，优化算法为 ADAM。Dropout 率为 0.1。预设有两个 GCN 编码器单元。前者维度数 E 为 64，后者维度数 E' 为 32。

图 5 是模型在三组数据集中训练集 loss 值的变化情况。实验结果表明模型在迭代数 epochs 达 200 时稳定收敛，因此选定模型参数 epochs=200。

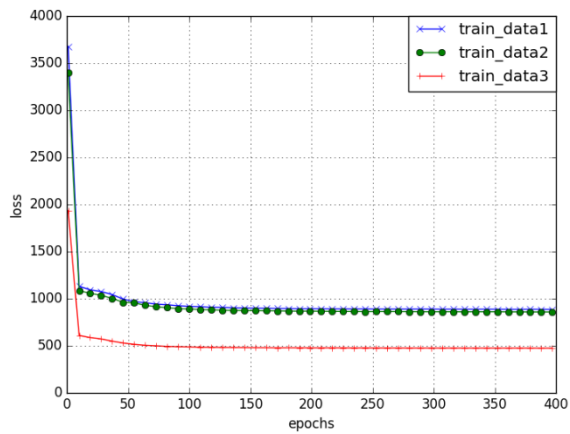

图 5 三组数据训练 loss 值的变化

图 6 表示三组数据随着训练迭代次数的增加，RMSE 值的变化情况。图中 3 条曲线变化的趋势是一致的。由图观察到数据集小的（data1）训练过程存在一个小幅度的波动。同时也说明即使在不同时间跨度不同数量级的训练数据，模型都能学习数据的特征。

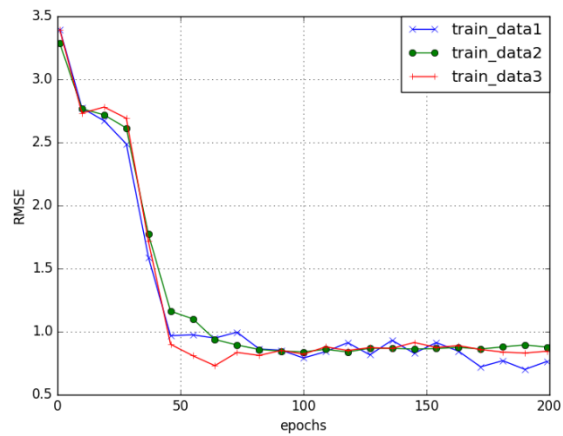

图 6 三组数据训练 RMSE 变化

图 7 表示使用 data1 训练模型得到的测试集评估指标。随机实验 10 次得到的各项评估指标。由于神经网络模型有很多训练参数，我们一般采用随机赋值网络中的初始参数变量，每次训练随机产生的初始变量值存在差异，因此模型训练的 loss 值以及预测结果也存在一个稳定范围内的波动，我们采用随机实验取均值的方式表示模型的训练效果。

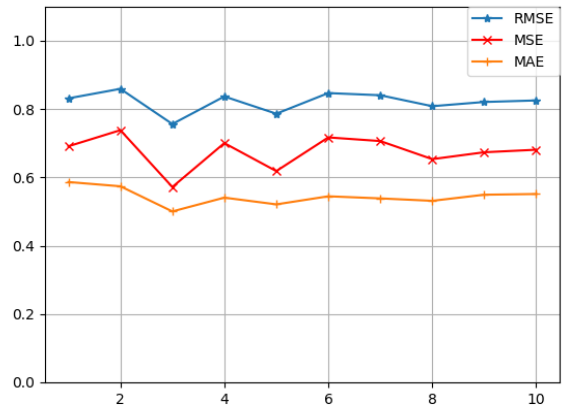

图 7 模型 10 次训练测试集的三个指标

4.3 对比试验

我们进行了两方面的对比实验，一是将我们的模型（Graph-VAE）和 GCMC 进行对比；二是我们调用开源实现 LibRec 包中三组共 13 中代表性的预测算法（分类聚类成绩预测、矩阵分解成绩预测以及协同过滤成绩预测）与 Graph-VAE 对比。

4.3.1 Graph-VAE 和 GCMC 对比

我们用同样的输入数据训练模型 Graph-VAE 和 GCMC,根据评价指标 RESM 如图 8 所示，Graph-VAE 效果要优于 GCMC，其中 ecoph 小于 50 时，Graph-VAE 下降较慢正是因为我们使用高斯混合模型刻画了学生与课程之间的相关性及差异性，防止过拟合。

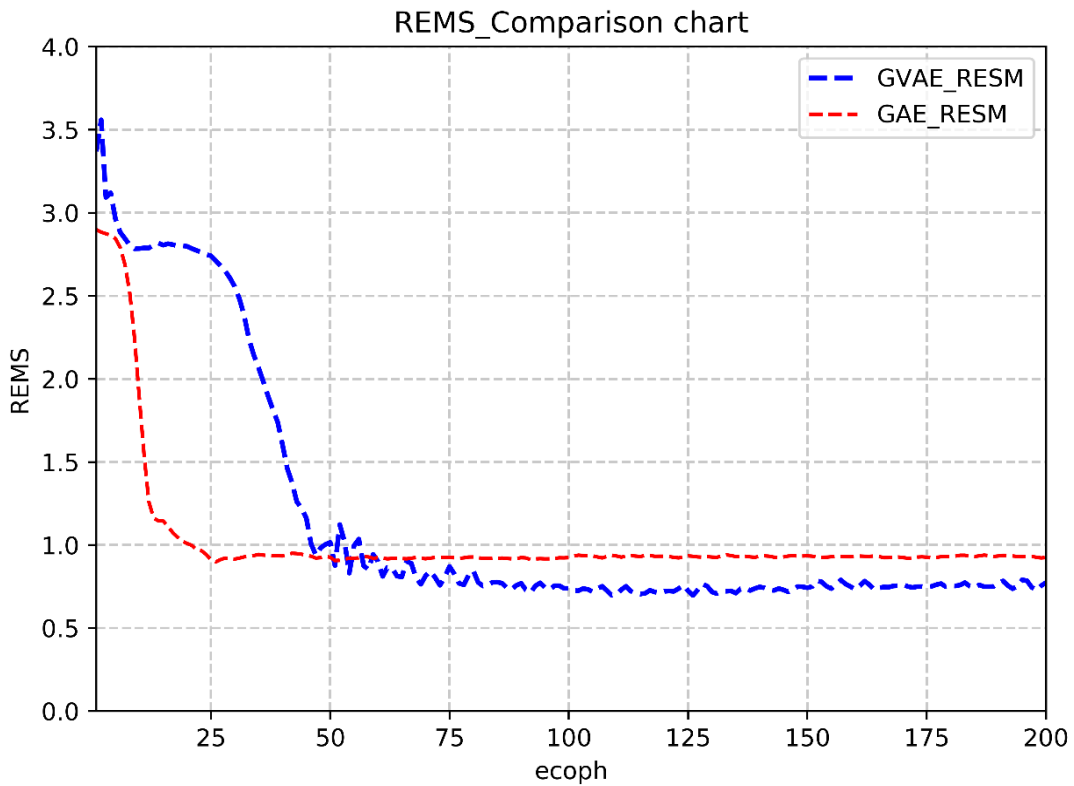

图 8 Graph-VAE 和 GCMC 的 RESM 对比图

### 4.3.2 Graph-VAE 与推荐算法比较

典型的推荐算法大致分为分类、聚类成绩预测、矩阵分解成绩预测以及协同过滤成绩预测三类。其中，分类、聚类成绩预测主要包含基于学生用户邻域推荐的 k 近邻算法（userknn）、基于课程邻域推荐的 k 近邻算法（itemknn）、基于学生用户推荐的聚类算法（usercluster）、基于课程推荐的聚类算法（itemcluster）等 4 种方法；矩阵分解方法成绩预测主要包含奇异值分解++（svdpp）、非负矩阵分解[21]（nmf）、biased 矩阵分解（biasedmf）、局部低秩张量矩阵分解算法（llorma）以及贝叶斯概率矩阵分解算法[22]（bpmf）等 5 种方法；协同过滤成绩预测[8]主要包含基于全局的协同过滤推荐算法（globalaverage）、基于内容的推荐算法

（aspectmodelrating）、基于课程的协同过滤推荐（Itemaverage）、基于学生的协同过滤推荐（useraverage）等 4 种方法。

我们实验对比了以上 13 种方法，如图 9 所示 Graph-VAE 的 RMSE 明显小于其他推荐算法，验证了模型的有效性。

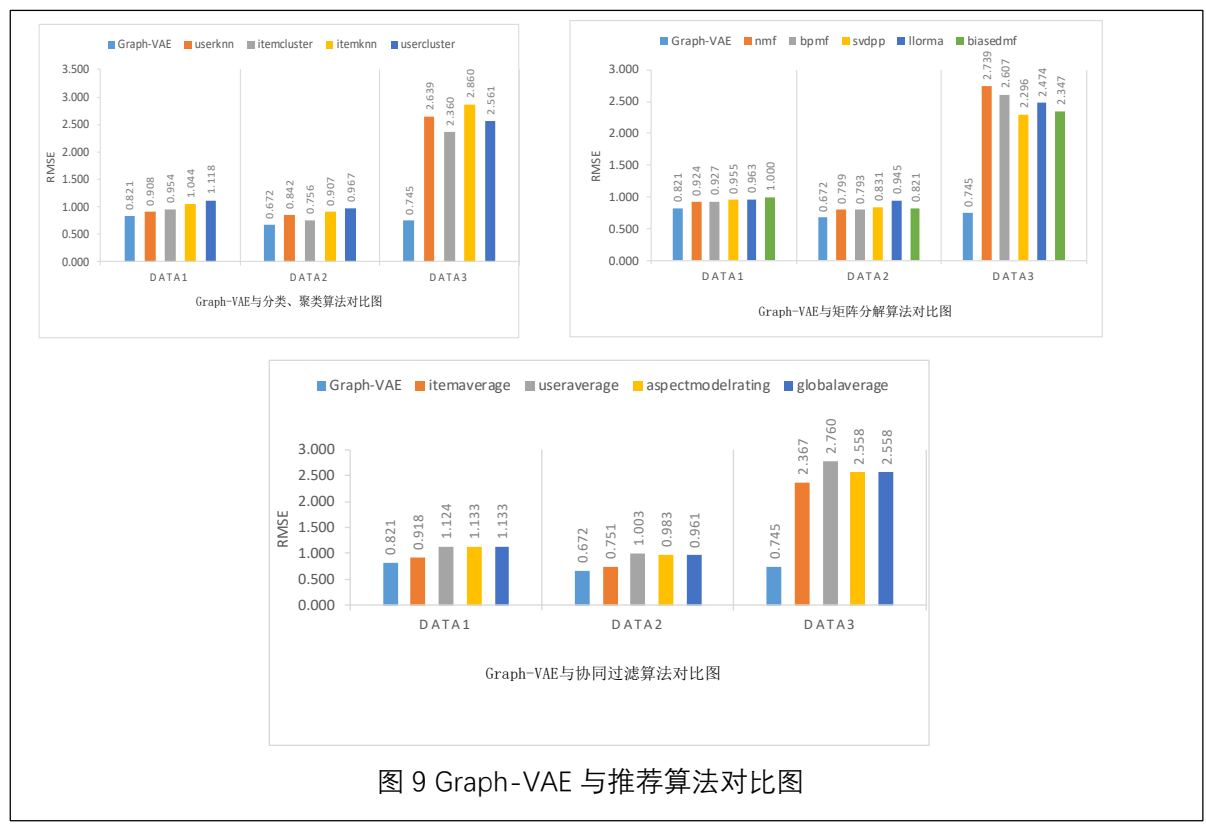

图 9 Graph-VAE 与推荐算法对比图

## 4.4 可视化分析

t-SNE（t-Distributed Stochastic Neighbor Embedding）[18]是一种应用广泛的用于非线性降维的算法，适于对高维数据进行可视化。

使用 t-SNE 可视化 GCN 学习到的学生向量 $Z_1,...,Z_{N_u}$，图 10 左边表示聚类之前向量的分布，右边为迭代次数为 2746 次后向量的分布。图中两个红圈内部的点可以认为是相近的学生向量。

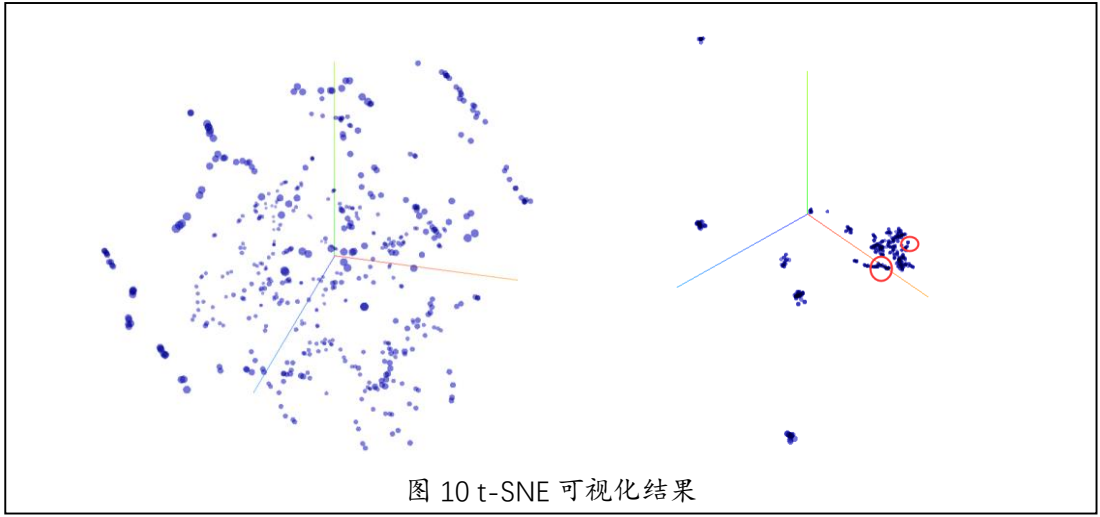

图 10 t-SNE 可视化结果

利用 t-sne 可视化节点向量可以将相近的节点聚到一起，根据聚类效果，实验选择了两个学生用户（图 10 右边两个红色圈上的点）来进行观察聚类在一起的学生的成绩分布。分别以选择的两个节点找与其相近的节点分成两组，图 11 为两个学生节点的特征向量聚类后相近节点成绩分布，可以看出两组学生各自每组的分布比较接近，尤其是高分分布和低分分布的规律比较明显。

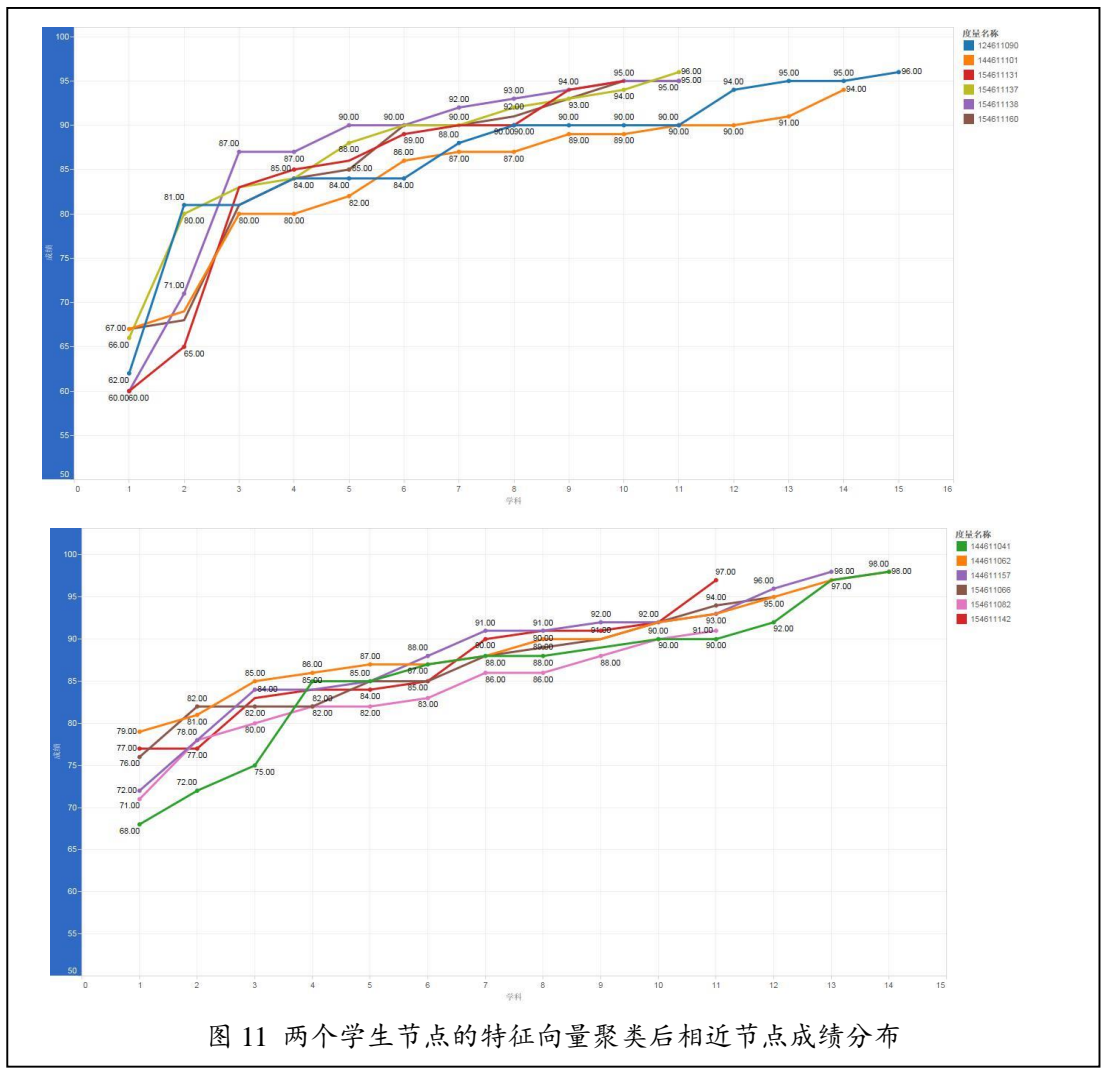

图 11 两个学生节点的特征向量聚类后相近节点成绩分布

4.5 归因分析

通过基于梯度的归因方法 DeepExplain[23]分析可知影响成绩预测的因素。通过分析可以知道课程之间的相互影响。此部分，我们共进行两种分析，一是分析学生节点所选课程（节点的相邻节点）中对成绩预测影响较大的节点；二是分析所有与学生节点相关的节点（节点的二跳邻居）对成绩预测影响较大的节点，企图发现学生之间发关系。

通过梯度归因方法对学生节点 ID 为 261 的相邻节点进行分析，研究该学生所选课程中对成绩预测影响较大的因素，可视化结果如图 12 所示。

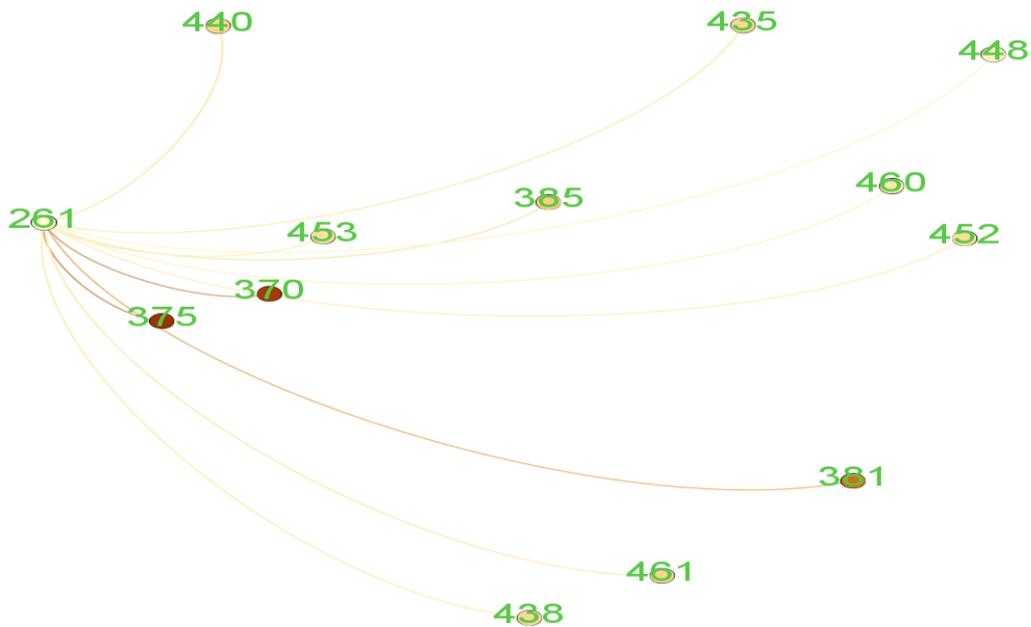

图 12 影响学生节点 261 成绩预测的因素。其中，其他节点是节点 261 的邻居节点颜色由浅到深表示表示节点对成绩预测影响程度的由低到高。颜色较深的为 370、375 和 381，它们分别对应课程自然辨证法、中国特色社会主义以及矩阵论三门公共课。

对于学生节点 261 来说，对成绩预测影响较大的是公共课成绩。经过调查发现，学生节点 261 是研究生，而该学校的研究生课程中，专业课考试相较于公共课简单，成绩普遍偏高，学生之间差异性较少，但是公共课考试困难，要学生花很大的精力学习和复习，学生成绩差异性较大。公共课考试成绩较好的一般都是上课认真听课，课下认真复习的学习态度好的学生，而专业课的高分成绩不能说明该学生学习认真。

通过归因分析方法对影响节点 ID 为 326_479(ID=326 的学生，所选课程的 ID=479，节点 479 对应课程机器学习与数据挖掘)的成绩预测的所有因素进行分析，研究影响对该同学该课程影响较大的因素是哪些，结果如图 13 所示。

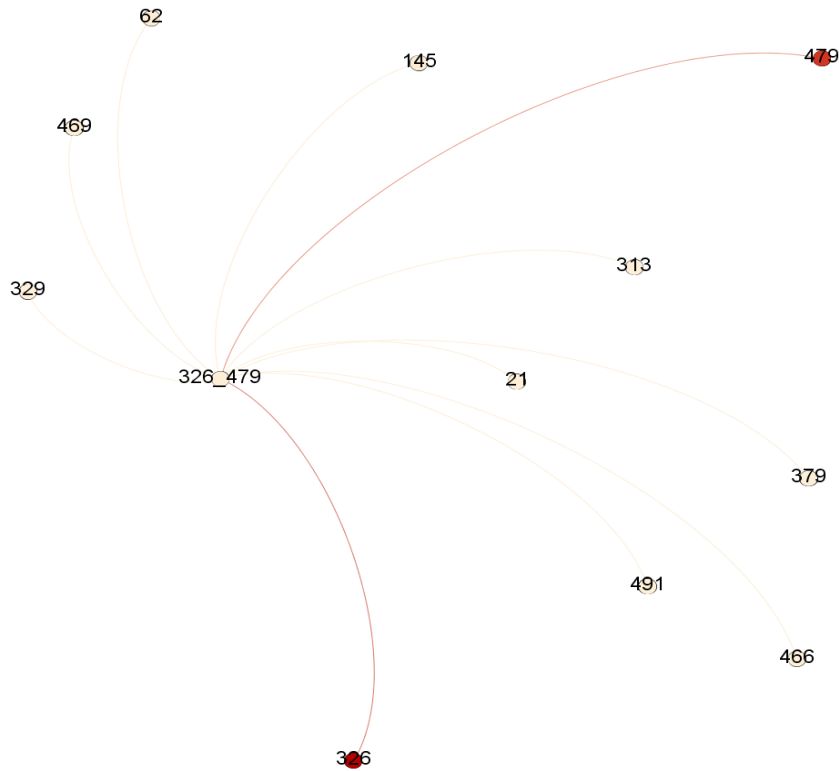

图 13 影响节点 ID 为 326_479 成绩预测较大的前 12 个节点。其中，节点 326 和节点 479 影响最大，属于自身对节点成绩预测的影响，紧跟其后分别有两种节点：①学生节点，分别是节点 145、62 和 21；②课程节点，分别是节点 469、379 和 466，其对应课程名称为专题研究、应用统计和高等计算机算法。

对于节点 326 的 479 这门课程的成绩预测，通过对所有结点进行分析，发现影响较大的是因素分别有学生节点和课程节点。调查研究发现，模型 Graph-VAE 对该学生该成绩预测为 77 分，而影响较大的几个学生成绩均在 70~80 之间，学生成绩分布相似。影响较大的课程分别是专题研究、应用统计和高等计算机算法。前两门课程均是公共课，符合第一个实验的说明，而高等计算机算法课程是机器学习与数据挖掘课程的基础课程，机器学习与数据挖掘课程中设计很多算法问题，也就是说高等计算机算法学习较好的同学相对来说更容易接受机器学习与数据挖掘课程所学知识，这符合现实基础课程对上层课程影响的现象，由此可以对学生选课提供帮助。

## 5 Conclusion

成绩预测是对学生个体进行建模并针对性改进教学的手段。针对学生学习过程中存在的问题，我们应该早发现，早预防。教育大数据的涌现使得深度学习应用于成绩预测成为可能，深度学习自动提取特征的结构尤其适于挖掘大样本数据的隐藏模式。本文提出了 Graph-VAE 模型，这种模型相比以往模型能自动抽取学生成绩数据的关键特征，从而完成成绩预测任务，促进教育教学。此外通过梯度归因分析，挖掘学生成绩数据中隐含的信息，为学生选课提供帮助。本文是针对此类问题的一次积极尝试。